\documentclass[10pt, conference]{ieeeconf} 
\IEEEoverridecommandlockouts                             

\usepackage{amssymb}

\usepackage{subfig}
\usepackage{multicol}
\usepackage[bookmarks=true]{hyperref}
\usepackage{graphicx,color,subfig}
\usepackage{amsfonts, amssymb, amsmath,verbatim, multirow}
\usepackage{enumerate}
\usepackage{breqn}
\usepackage{amsmath}
\usepackage{diagbox}
\usepackage{balance}
\usepackage{times}
\usepackage{algorithm}
\usepackage{algorithmic}
\usepackage{easybmat}

\usepackage{caption}

\newcommand{\mb}[1]{{\boldsymbol{#1}}}

\newcommand{\todo}[1]{\textcolor{blue}{#1}}

\usepackage[usenames,dvipsnames]{xcolor}
\usepackage{verbatim}  
\title{\Large \bf Motion2Vec: Semi-Supervised Representation Learning from Surgical Videos}
\author{Ajay Kumar Tanwani$^{1}$, Pierre Sermanet$^{2}$, Andy Yan$^{1}$, Raghav Anand$^{1}$, Mariano Phielipp$^{3}$, Ken Goldberg$^{1}$
\thanks{$^{1}$ University of California, Berkeley. \newline \text{\{ajay.tanwani, yan.andy4, raghav98, goldberg\}@berkeley.edu}}
\thanks{$^{2}$ Google Brain, USA. \text{sermanet@google.com}}
\thanks{$^{3}$ Intel AI Labs, USA. \text{mariano.j.phielipp@intel.com}}
}

\begin{document}
\maketitle
\begin{abstract}
Learning meaningful visual representations in an embedding space can facilitate generalization in downstream tasks such as action segmentation and imitation. In this paper, we learn a motion-centric representation of surgical video demonstrations by grouping them into action segments/sub-goals/options in a semi-supervised manner. We present Motion2Vec, an algorithm that learns a deep embedding feature space from video observations by minimizing a metric learning loss in a Siamese network: images from the same action segment are pulled together while pushed away from randomly sampled images of other segments, while respecting the temporal ordering of the images. The embeddings are iteratively segmented with a recurrent neural network for a given parametrization of the embedding space after pre-training the Siamese network. We only use a small set of labeled video segments to semantically align the embedding space and assign pseudo-labels to the remaining unlabeled data by inference on the learned model parameters. We demonstrate the use of this representation to imitate surgical suturing motions from publicly available videos of the JIGSAWS dataset. Results give $85.5$\% segmentation accuracy on average suggesting performance improvement over several state-of-the-art baselines, while kinematic pose imitation gives $0.94$ centimeter error in position per observation on the test set. Videos, code and data are available at: \todo{\url{https://sites.google.com/view/motion2vec}}
\end{abstract}

\section{Introduction}

A long-standing goal in artificial intelligence is to learn new skills by observing humans. Learning manipulation skills from video demonstrations by imitation can provide a scalable alternative to traditional kinesthetic and teleoperation interfaces. Generalizing these skills to new situations requires extracting disentangled representations from observations such as the relationships between objects and the environment while being invariant to lighting, background, and other geometric properties such as position, size of external objects and viewpoint of the camera.

\begin{figure}[!tbp]
\centering
\includegraphics[trim={0.0cm 0.cm 0.0cm 0.cm},clip,scale = 0.451]{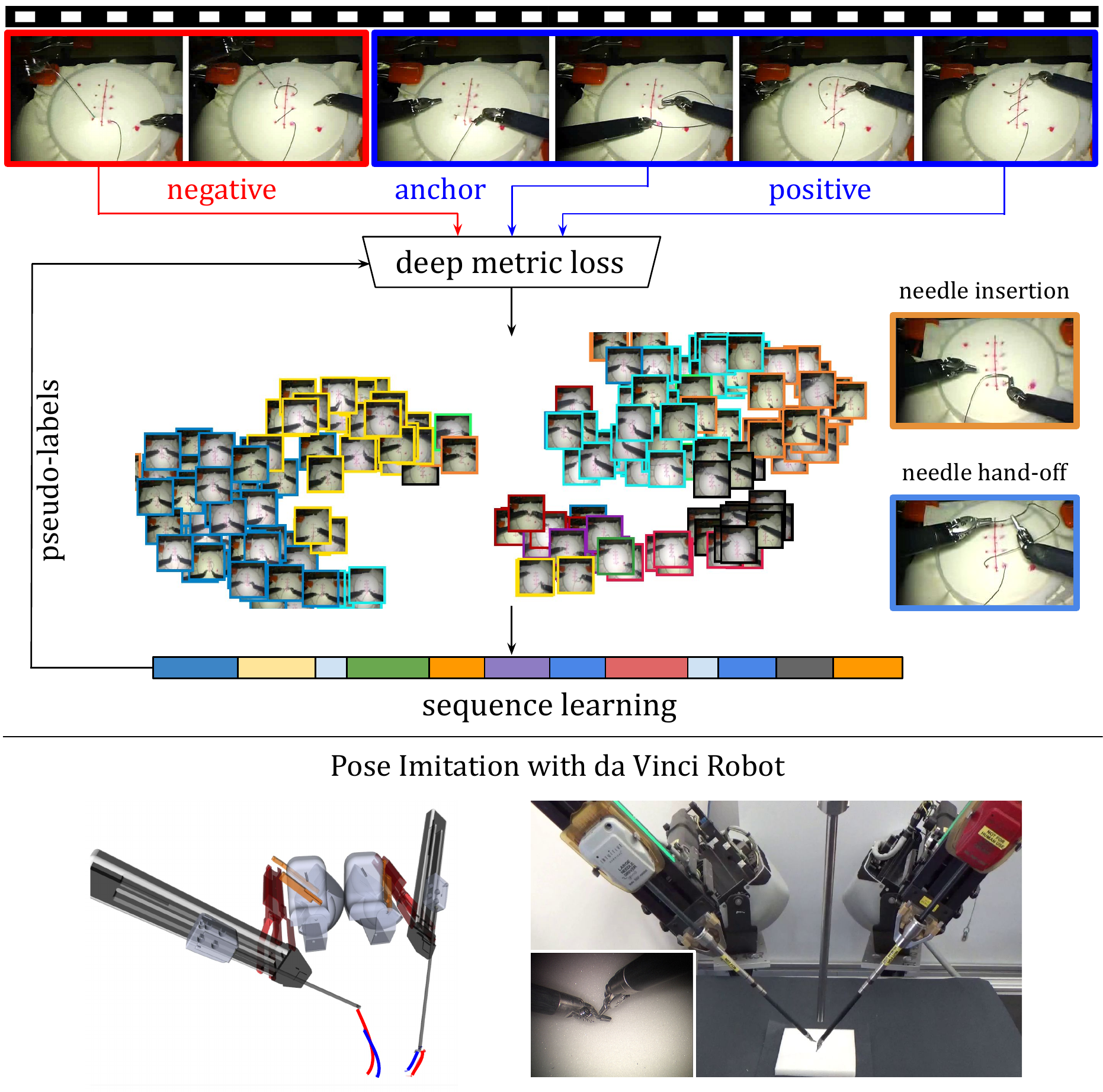}
\caption{\footnotesize Motion2Vec groups similar action segments together in an embedding space in a semi-supervised manner. The embedding space is represented above with a t-SNE plot. Pseudo-labels for the unlabeled training data are obtained by inference on a RNN that is trained iteratively for a given parametrization of the Siamese network. The learned representation is applied to surgical suturing segmentation and pose imitation in simulation and real on the da Vinci robot.} \label{fig: m2v_cover}
\end{figure}

When the demonstrations have sequential, recursive, relational or other kinds of structure, structured representations can be useful to infer meaningful hidden associations. *2Vec models such as Word2Vec~\cite{Tomas_NIPS_13}, Grasp2Vec~\cite{Jang_grasp2vec_18}, and Demo2Vec~\cite{Fang18} capture such data-centric relationships by bringing similar observations together in an embedding space. In this paper, we present Motion2Vec, an algorithm that acquires motion-centric representations of manipulation skills from video demonstrations for imitation learning (see Fig.~\ref{fig: m2v_cover}). Consistency, interpretability and supervisory burden are a key concern in imitation learning as it is often difficult to precisely characterize what defines a segment and labeling can be  time-consuming. We seek to encode the observations with weak supervision using a small set of labeled videos, while allowing better generalization and interpretability for new situations. Consider, for example, the surgical suturing task that may be decomposed into temporally connected movement primitives or action segments like \textit{needle insertion}, \textit{needle extraction}, \textit{needle hand-off} and so on. Such a hierarchical decomposition is analogous to the speech recognition and synthesis problem where words and sentences are synthesized from phoneme and triphone based segments~\cite{Rabiner89}. Other related domains include activity recognition in computer vision~\cite{Yeung16,Zhou18}, options in hierarchical reinforcement learning~\cite{Stolle02} and parts of speech tagging in natural language processing \cite{Kupiec92}.


The focus of the paper is on extracting disentangled representations from video demonstrations in a semi-supervised manner for downstream tasks of action segmentation and imitation. We use metric learning with a Siamese network to bring similar action segments -- images with same discrete labels -- together in an embedding space. After pre-training the network, we use a recurrent neural network to predict pseudo-labels on unlabeled embedded sequences that are fed back to the Siamese network to improve the alignment of the action segments. Motion2Vec moves the video observations into a vector domain where closeness refers to spatio-temporal grouping of the same action segments.


We evaluate its application to segment and imitate surgical suturing motions on the dual arm da Vinci robot from publicly available videos of the JIGSAWS dataset. We compare the proposed approach with several state-of-the-art metric and sequence learning methods including temporal cycle-consistency (TCC)~\cite{Dwibedi_19}, single-view time contrastive network (svTCN)~\cite{Sermanet17}, temporal convolutional network (TCN)~\cite{Lea_16}, hidden Markov model (HMM)~\cite{Rabiner89}, hidden semi-Markov model (HSMM)~\cite{Yu10}, conditional random fields (CRF)~\cite{Lafferty01}, and obtain better segmentation accuracy of $85.5$\% on the leave one super trial out test set than reported in the literature, e.g.,~\cite{DiPietro_16,Lea_16} report accuracy of $83.3$\% and $81.4$\% respectively. Moreover, we introduce pose imitation results on the da Vinci robot arms with $0.94$ centimeter error in position per observation respectively.

\subsection{Contributions}
This paper makes the following contributions:
\begin{enumerate}
    \item A novel Motion2Vec representation learning approach for spatio-temporal alignment of action segments/sub-goals/options in an embedding space using a small set of labeled video observations in a semi-supervised manner.
    
    \item Performance analysis with different combinations of supervised and unsupervised approaches to metric and sequence learning for extracting meaningful segments from surgical videos.
    
    
    
    
    
    \item Learning surgical suturing segments and kinematic imitation of poses on da Vinci robot arms from publicly available videos of the JIGSAWS dataset suggesting performance improvement over several state-of-the-art methods.
\end{enumerate}

\section{Related Work}
\textbf{Imitation learning} provides a promising approach to teach new robotic manipulation skills from expert demonstrations. Common approaches to imitation learning include \textit{behaviour cloning} and \textit{inverse reinforcement learning (IRL)} with Dynamic Movement Primitives \cite{Ijspeert13}, Gaussian mixture models~\cite{Calinon17}, task-parametrized generative models~\cite{Tanwani18}, Generative Adversarial Imitation Learning \cite{Ho16}, one-shot imitation learning \cite{Duan17}, Dagger~\cite{Ross10}, and behavior cloning from observation~\cite{Torabi_18} (see~\cite{Osa18} for an overview). In contrast to direct trajectory learning approaches from demonstrations~\cite{Argall09}, we focus on self-supervised and semi-supervised approaches to learn skills from video observations where only a few segment labels may be available.

\textbf{Weakly Supervised Learning from Videos:} Uncontrolled variables such as lighting, background and camera viewpoint pose a challenge to robot learning from video observations. Learning from multiple viewpoints, temporal sequences, labeled action segments, weakly supervised signals such as order of sub-actions, text-based annotations or unsupervised learning are feasible alternatives to learning dense pixel-wise visual descriptors from observations~\cite{zeng_3dmatch_2016,Schmidt_RAL_17,Florence_dod_18}. Kuehne et al. presented a generative framework for end-to-end action recognition by extracting Fisher vectors from videos and sequencing them with HMMs~\cite{Kuehne15}, followed by a weakly supervised approach for temporal action segmentation with RNN-HMM~\cite{Kuehne19}. Tang et al. learn temporal structure in the videos for complex event detection~\cite{Tang_cvpr_2012}. Liu et al. learn a translation invariant policy between the expert and the learner contexts~\cite{Liu17}. Doersch et al. use spatial coherence in the neighbouring pixels for learning unsupervised visual representations~\cite{Doersch15}. Misra et al. propose shuffle and learn to maintain temporal order in learning the visual representation~\cite{Misra16}. Wang and Gupta use triplet loss to encourage the first and the last frame of the same video together in the embedding space, while pushing away the negative sample from another class~\cite{Wang15a}. Sermanet et al.~\cite{Sermanet17} use metric learning loss for getting a temporally coherent viewpoint invariant embedding space with multi-view or single view images and use nearest neighbours in the embedding space for imitation learning. Dwibedi et al. extend the approach to multiple frames in ~\cite{Dwibedi_18}, and use a temporal cycle consistency (TCC) loss by matching frames over time across videos in~\cite{Dwibedi_19}. Finn et al. present a deep action-conditioned visual foresight model with model-predictive control for learning from pixels directly~\cite{Finn_VF_16,Ebert_18}. 

Siamese networks learn a similarity function across images in an embedding space~\cite{Schroff15,Koch2015,Baraldi_15,Hermans17,Roy_18}. In this paper, we encode the video observations based on spatio-temporal alignment of action segments/options, in contrast to using time only with self-supervised representations~\cite{Sermanet17}. We use triplet loss to attract similar action segments in an embedding space and repel samples from other action segments in a \textbf{semi-supervised} manner. We only use a small set of labeled video segments to semantically align the embedding space and predict pseudo-labels for the remaining unlabeled data by iteratively segmenting and learning the embedding space with a hybrid DNN-RNN model. 

\textbf{Surgical Suturing: }Surgical suturing automation has been studied in several contexts such as needle path planning~\cite{Sen_16}, collaborative human-robot suturing~\cite{Padoy_11}, and learning from demonstrations by trajectory transfer via non-rigid registration in simulation~\cite{Schulman_13}. The suturing motions are decomposable into simpler sub-tasks or surgemes that can be inferred from demonstrations~\cite{Lea_15,Murali16,Krishnan18}. In this work, we apply the Motion2Vec approach to infer action segments/surgemes and end-effector poses on the dual arm da Vinci robot from publicly available suturing videos of the JIGSAWS dataset~\cite{gao14}. Results suggest performance improvement in segmentation over state-of-the-art baselines~\cite{DiPietro_16,Ahmidi_17}, while introducing pose imitation on this dataset with $0.94$ cm error in position per observation respectively.

\section{Motion2Vec Learning from Videos}
%

\subsection{Problem Formulation}\label{sec: PS}
Consider $\{\mb{I}_{n,t}\}_{n=1, t=1}^{N, T_n}$ as a set of $N$ video demonstrations, where $\mb{I}_{n,t} \in \mathbb{R}^{640 \times 480 \times 3}$ denotes the RGB image at time $t$ of $n$-th demonstration comprising of $T_n$ datapoints. Each demonstration describes a manipulation skill such as pick-and-place or surgical suturing task, collected from a third-person viewpoint that does not change in a demonstration but may change across demonstrations. We assume access to a supervisor that assigns segment labels to small set of the demonstrations as belonging to one of the segments $z_{m,t} \in \{1 \ldots C \}$ such as reach, grasp etc., resulting in a set of labeled demonstrations $\{\mb{I}_{m,t}, z_{m,t}\}_{m=1, t=1}^{M, T_m}$ with $M \ll N$. Without loss of generality, we drop the indices $n$ and $m$ to denote an image frame as $\mb{I}_t$ for the rest of the paper. 

We seek to learn a deep motion-centric representation of video observations $f_{\mb{\Theta}_{\text{D}}} : \mb{I}_t \rightarrow \mb{\xi}_t$ with $\mb{\xi}_t \in \mathbb{R}^{d}$ and $d \ll \vert \mb{I}_t\vert$ such that similar action segments are grouped together in the embedding space while being invariant to the nuisance variables such as lighting, background and camera viewpoint. The representation needs to semantically align the unlabeled demonstrations with a small set of labeled demonstrations in a semi-supervised manner. We predict pseudo-labels $\mb{\hat{z}}_{t-l:t}$ for the unlabeled training mini-batch of length $l$ with a sequence learning model $h_{\mb{\Theta}_{\text{S}}} : \mb{\xi}_{t-l : t} \rightarrow \mb{\hat{z}}_{t-l:t}$ and feed them back to the embedding network to align the action segments. After pre-training the embedding network, the pseudo-labels and the deep embedding space are alternatively updated for a given estimation of the sequencing model $\mb{\hat{\Theta}}_{\text{S}}$ and the deep embedding space $\mb{\hat{\Theta}}_{\text{D}}$ in an iterative manner, i.e.,
\begin{align}
\mb{\xi}_t \; = \; & f(\mb{I}_t; \mb{\Theta}_{\text{D}}, \mb{\hat{\Theta}}_{\text{S}}) \newline \\
\mb{\hat{z}}_{t-l:t} \; = \; & h(\mb{\xi}_{t-l:t}; \mb{\hat{\Theta}}_{\text{D}}, \mb{\Theta}_{\text{S}})
\end{align} 

The learned networks are subsequently used to train the control policy in the embedding space $\pi_{\mb{\Theta}_R}: \mb{\xi}_t \rightarrow \mb{u}_t$ where $\mb{u}_t \in \mathbb{R}^{p}$ corresponds to the end-effector pose of the robot arm to imitate the manipulation skill in the video demonstrations. In this work, we assume access to the kinematic poses of the end-effector in the video demonstrations, and defer autonomous learning of the manipulation skill from the embedded observations to the future work.

We follow a three-step alternating methodology to learn the Motion2Vec representation (see Fig.~\ref{fig: m2v_cover}): 1) learn the deep embedding space to pull together similar segments close while push away other far segments with a metric learning loss, 2) train the sequence model parameters using the labeled embedded observations, and 3) infer the most likely segments for the video observations as pseudo-labels for the unlabeled training data to refine the embedding space.

\subsection{Deep Embedding Space}\label{sec: ES}
The deep embedding space reflects the task relevant attributes of the objects in the videos and how they can be mapped onto the robot end-effector. Sample and time complexity of collecting/labelling videos at pixel level to reflect such associations can be very high for training vision-based deep models in robotics. The trajectory-centric invariant formulations as in \cite{Ijspeert13,Khansari11,Tanwani_WAFR_18} may not readily generalize to the video demonstrations. First, transforming the images with respect to an arbitrary viewpoint is non-trivial as it requires the full 3D reconstruction of the environment. Second, the variance across images per pixel may not be indicative of the representative features in the demonstrations. 

In this work, we learn a deep embedding representation using a metric learning loss which pulls together observations from the same action segment in the embedding space, while pushing away observations from other action segments that functionally correspond to different sub-goals or movement primitives. We use triplet loss for metric learning in this work~\cite{Schroff15}. Note that the contrastive loss or the magnet loss may also be used in a similar way~\cite{Rippel15}. During training, the loss operates on the tuple corresponding to the anchor image embedding $\mb{\xi}_t$, a positive sample belonging to the same action segment $\mb{\xi}_t^{+}$ and a negative sample randomly chosen from another action segment $\mb{\xi}_t^{-}$. The segment labels for the unlabeled training data are predicted by inference on the current estimate of the sequence model parameters (see Sec.~\ref{sec: seq_learn}). Triplet loss posits that the distance of the anchor to the positive sample in the embedding space is less than the distance to the negative sample by some constant margin $\zeta$, i.e.,
\begin{equation}
\mathcal{L}(\mb{\Theta}_{\text{D}}, \mb{\hat{\Theta}}_{\text{S}}) = \frac{1}{T} \sum_{t=1}^{T} \left\{ \Vert \mb{\xi}_t \; - \; \mb{\xi}_t^{+} \Vert_2^{2} - \Vert \mb{\xi}_t - \mb{\xi}_t^{-} \Vert_2^{2} \; + \; \zeta \right\}_+,
\end{equation}
where $\{.\}_+$ is the hinge loss and the representation $\mb{\xi}_t$ is normalized to extract scale-invariant features similar to~\cite{Schroff15}. We compare the triplet metric learning loss with other embedding approaches including: 1) \textbf{incremental principal component analysis (iPCA)} to project video observations into an uncorrelated embedding space~\cite{Zhao_06}, 2) \textbf{temporal cycle consistency (TCC)} to align the embedding space by matching frames over time across video demonstrations~\cite{Dwibedi_19}, 3) \textbf{time-contrastive network (svTCN)} with single view using a window of $6$ neighbouring frames in the sequence to find the positive sample for each anchor image and negative sample from a window of $12$ neighbouring frames~\cite{Sermanet17}, 4) \textbf{N-pairs} metric that takes pairs of images from same segment labels where a pair is used as an anchor and a positive image respectively, while each pair in the mini-batch may have different labels. The n-pairs loss repels a positive sample from all negative samples in comparison to the nearest negative sample in triplet loss~\cite{Sohn_16}. Note that iPCA, TCC and svTCN are used in an unsupervised way, whereas we use triplet loss and n-pairs loss in a (semi-)supervised manner.

\subsection{Sequence Learning and Inference of Action Segments}\label{sec: seq_learn}
We capture the spatio-temporal dependencies in the embedded observations to predict action segments with a sequence learning model. We use a \textbf{recurrent neural network (RNN)} to discriminatively model the action assignment to the observation sequence $P(\mb{\hat{z}}_{t-l:t} \; \vert \; \mb{\xi}_{t-l:t})$ using a stride of length $l$ in a mini-batch. A RNN maintains an additional hidden state and uses the previous hidden state and the current input $\mb{\xi}_{t}$ to produce a new hidden state and the output $\hat{z}_t$. The hidden state preserves the effect of previous observations in predicting the current output. We use the bi-directional Long Short-Term Memory (LSTMs)~\cite{Hochreiter_97} in this work that can also preserve the effect of future observations within a sequence. We minimize the cross-entropy loss between the true and the predicted labels during training with backpropogration through time~\cite{Graves_12}. We infer the most likely sequence of action segment labels on the unlabeled training data after pre-training the network and only keep the top-$k$ pseudo-labeled examples for each action segment. The psuedo-labels have the same effect as that of entropy regularization that pushes the decision boundaries to be in well-separated low density regions~\cite{Lee_13}. 

We compare RNNs with \textbf{$k$-nearest neighbours (KNN)} classification accuracy in the embedding space, along with other sequence learning models, namely: 1) \textbf{Conditional Random Fields (CRFs)} that encode the conditional probability distribution of the output labels sequence given the input observation sequence in a discriminative manner~\cite{Lafferty01,Sutton12,Vakanski12}, 2) \textbf{Hidden Markov Models (HMMs)} that augment the deep embedding space with latent states that sequentially evolve over time in the embeddings \cite{Rabiner89}, 3) \textbf{Hidden Semi-Markov Models (HSMMs)} that relax the Markovian structure of state transitions in a HMM by relying not only upon the current state but also on the duration/elapsed time in the current state, i.e., the underlying process is defined by a \textit{semi-Markov} chain with a variable duration time for each state~\cite{Yu10,Tanwani16}. Note that CRFs and RNNs are discriminative models trained in a supervised manner, while HMM and HSMMs are generative models used in an unsupervised way (see supplementary materials for more details of the compared approaches).

The sequence labels $\mb{\hat{z}}_{1:T}$ on unlabeled data are estimated for a given set of embedding and sequence model parameters, and further used to update the parameters of the deep embedding space. Step~\ref{sec: ES} and  Step~\ref{sec: seq_learn} are repeated until convergence.

\subsection{Imitating End-Effector Poses} \label{sec: samplingHSMM}

Given the learned model parameters of the embedding network, we map the embedded observation $\mb{\xi}_t$ to the end-effector pose $\mb{u}_t$ of the robot arm by behaviour cloning with a feedforward neural network $\pi_{\mb{\Theta}_R}$. The feedforward network is defined on top of a pretrained embedding network whose parameters are frozen during the learning process. The pose imitation loss is a weighted combination of the position loss measured in terms of the mean-squared error between the ground-truth and the predicted end-effector position, and the orientation loss measured in terms of the cosine distance between the ground-truth and the predicted end-effector orientation in quaternion space.

\section{Experiments, Results and Discussions} \label{sec: Exp}
We evaluate Motion2Vec representation for imitating surgical suturing motions on the dual arm da Vinci robot from the publicly available JIGSAWS dataset~\cite{gao14}. Note that we do not model contact dynamics with the needle and the suturing phantom, and only imitate the suturing motions on the kinematic level. We empirically investigate: 1) what metric/sequence learning representations generalize better in terms of the segmentation accuracy, 2) the effect of relative proportion of labeled examples in semi-supervised learning, and 3) the usefulness of the learned embeddings in imitating the end-effector poses on the da Vinci arms.

\subsection{JIGSAWS Dataset}
JIGSAWS dataset contains video demonstrations of three surgical tasks, namely suturing, needle-passing and knot-tying. We only present results for imitating surgical suturing motions in this work. The suturing dataset consists of $8$ surgeons with varying skill levels performing the suturing demonstrations $5$ times each on the dual arm da Vinci robot. Each demonstration consists of a pair of videos from the stereo cameras, kinematic data of the end-effector of the robot arms, and the action segment label for each video frame among a distinct set of $11$ suturing sub-tasks as annotated by the experts. The discrete labels correspond to no activity stage \texttt{[IDLE]}, reach needle with right hand \texttt{[REACH-N-R]}, position needle \texttt{[POS-N]}, push needle through tissue \texttt{[PUSH-N-T]}, transfer needle from left to right \texttt{[TRANS-L-R]}, move to center \texttt{[MOVE-C]}, pulling suture with left hand \texttt{[PULL-L]}, orienting needle \texttt{[ORIENT-N]}, tighten suture with right hand \texttt{[TIGHT-R]}, loosening suture \texttt{[LOOSE-S]}, and dropping suture at the end \texttt{[DROP-S]}. The viewpoint of the camera, lighting and background is fixed in a demonstration, but changes slightly across demonstrations. The suturing style, however, is significantly different across each surgeon. We use a total of $78$ demonstrations from the suturing dataset downsampled at $3$ frames per second with an average duration of $3$ minutes per video. $62$ demonstrations with $4$ randomly chosen demonstrations from each surgeon are used for the training set ($1$ demonstration from a surgeon is corrupted and not used for training), while the remaining demonstration from all surgeons are used as the test set for a total of $16$ demonstrations. 
\subsection{Network Architecture(s)}
The Siamese network takes as input a downsampled $3$-channel RGB $320\times240$ image. The network is augmented on top of the Inception architecture, pre-trained on the ImageNet dataset. We add two convolutional layers of depth $512$ each on top of `Mixed-$5$d' layer followed by a spatial softmax layer~\cite{Finn_ss_15}, a fully connected layer of $2048$ neurons and an embedding layer of $32$ dimensions. We use the same Siamese network architecture in all the experiments. This embedding is trained on the triplet loss with a margin of $\zeta = 0.2$. We use a batch size of $128$ and $64$ for Siamese network and RNN respectively. Note that $64$ batch size at $3$ fps corresponds to online segmentation window of $21.3$ seconds.

The embedding sequence is fed to a $1$-layer bi-directional LSTM of $256$ hidden neurons. The CRFs network uses $32$ potential functions. The HMM/HSMMs are trained in an iterative manner with $K=30$ hidden states and a multivariate Gaussian in the observation distribution by pooling all the data from the embedding layer after every $1000$ iterations. The number of hidden states are empirically chosen between $1-50$ components to get best classification accuracy on the training set. The feedforward network for pose imitation consists of $6$ hidden units with $512$, $256$, $128$, $64$, $32$ and $16$ neurons. The output network corresponds to $3$-dimensional Cartesion position of the end-effector, $4$-dimensional quaternion orientation of the end-effector, and a jaw angle for each of the two arms.
\begin{table}[!tbp]
\caption{\footnotesize Segmentation accuracy performance comparison on the evaluation set averaged over $4$ iterations. Rows correspond to a different embedding space approach, columns correspond to a different segmentation method. KNN results are on training with a Siamese network only. CRF, RNN, N-pairs and Triplet models are trained in a supervised manner, while KNN, HMM, HSMM, PCA, TCC and svTCN are unsupervised. Motion2Vec (M2V) uses triplet loss with RNN, while M2V-T combines the triplet and svTCN loss for temporal alignment.} \normalsize \centering \label{tab: segment_comp}
\begin{tabular}{|c||c||c|c||c|c|}
\hline
 & KNN & HMM & HSMM & CRF & RNN \\ \hline \hline
\textbf{iPCA} & $0.586$  & $0.395$ & $0.392$ & $0.415$ & $0.721$ \\ \hline
\textbf{TCC} & $0.667$& $0.662$ & $0.642$ & $0.601$ & $0.727$ \\ \hline
\textbf{svTCN} & $0.792$& $0.676$ & $0.661$ & $0.723$ & $0.812$ \\ \hline \hline
\textbf{Images} & $0.748$& $0.716$ & $0.712$ & $0.811$ & $0.835$ \\ \hline
\textbf{N-Pairs} & $0.835$& $0.799$ & $0.794$ & $0.824$ & $0.854$ \\ \hline 
\textbf{M2V} & $0.829$& $0.831$ & $0.812$ & $0.838$ & $\textbf{0.855}$ \\ \hline
\textbf{M2V-T} & $\textbf{0.844}$& $0.828$ & $0.822$ & $0.801$ & $0.843$ \\ \hline
\end{tabular}
\end{table} 

\begin{figure}[tbp]
\centering
\includegraphics[trim={0.1cm 0.1cm 0.1cm 0.1cm},clip,scale = 0.15]{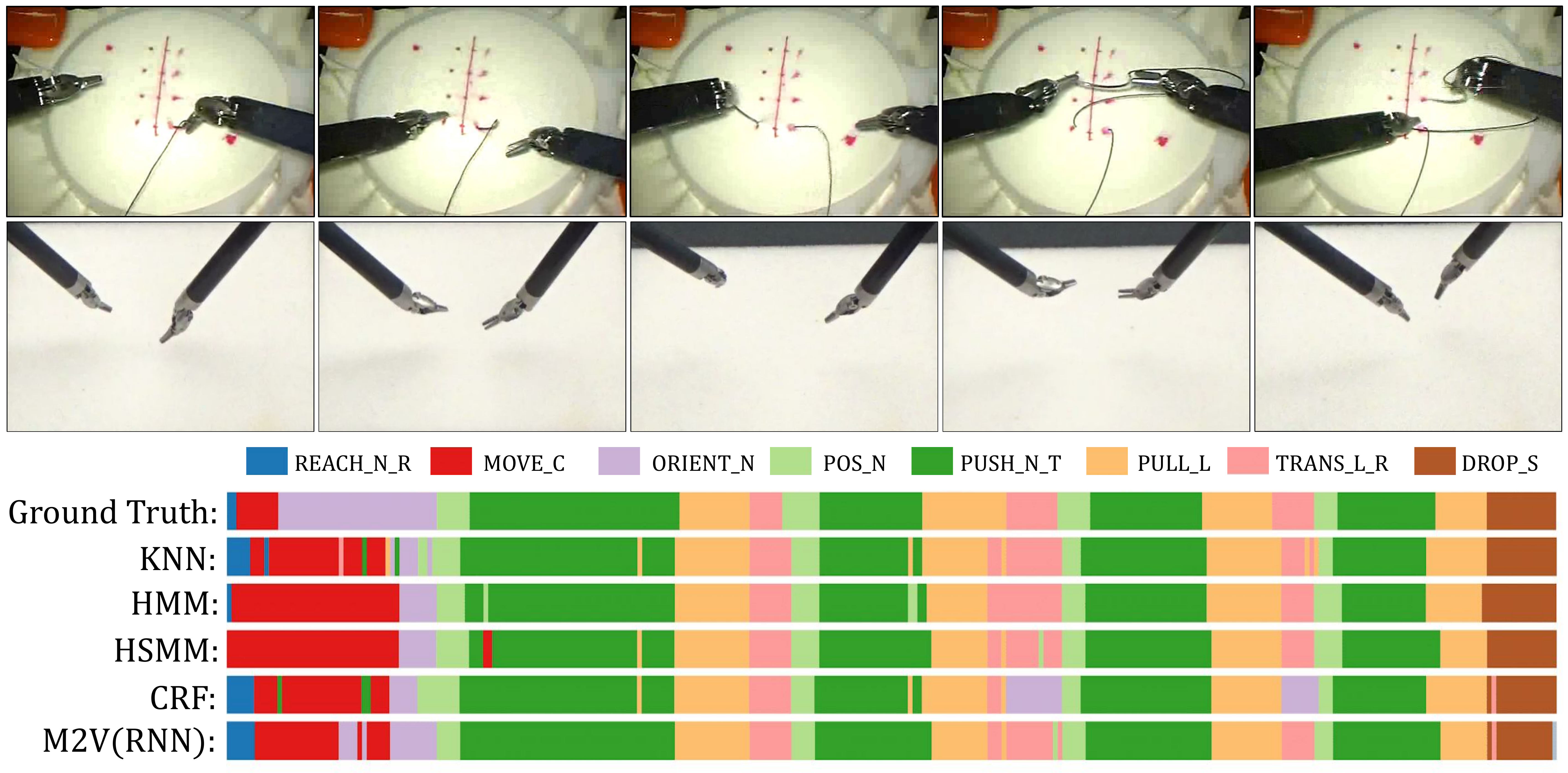}
\caption{\footnotesize \textit{(top)} Qualitative performance comparison of different segmentation policies on top of Motion2Vec with the ground-truth sequence for surgical suturing. Test set image sequence for the first suture is shown on \textit{(top)}, while the imitation sequence on da Vinci robot arms is shown on \textit{(bottom)}. KNN, HMM, HSMM are unsupervised, while CRF and RNN are supervised approaches. In comparison to KNN, RNN gives temporally consistent segments by using the sequential information in the embedded vectors.} \label{fig: seg-dnn-hsmm}
\end{figure}
\begin{figure}[!tbp]
\centering
\includegraphics[trim={0.1cm 0.1cm 1.0cm 0.4cm},clip,scale = 0.47]{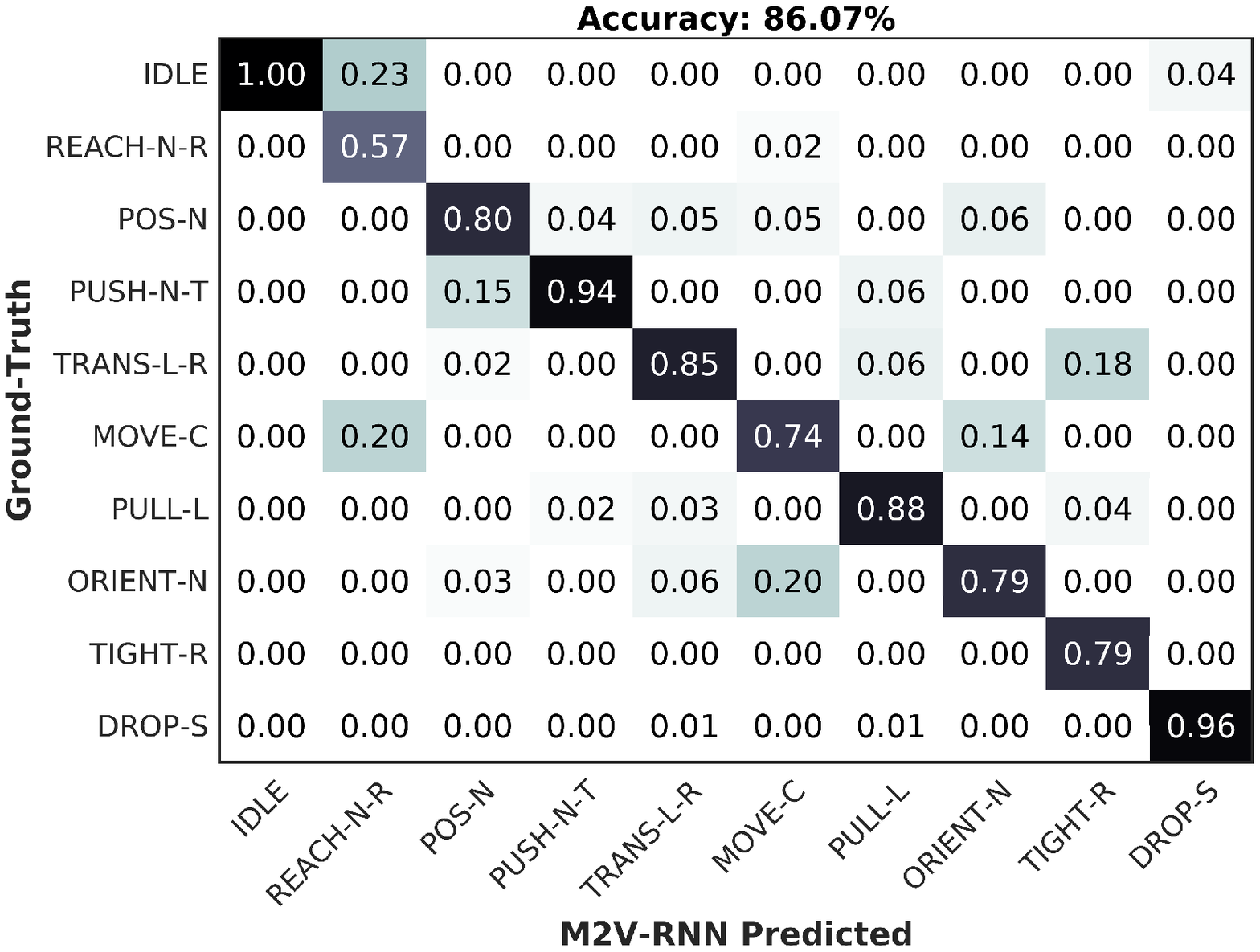}
\caption{\footnotesize Normalized confusion matrix of Motion2Vec with RNN on suturing test set, suggesting similar action segments are grouped together in the embedding space.} \label{fig: Conf_suturing}
\end{figure}
\subsection{Role of Supervision in Metric \& Sequence Learning}
Table~\ref{tab: segment_comp} summarizes the performance comparison of Motion2Vec with different combinations of supervised and unsupervised approaches to metric and sequence learning. We use segmentation accuracy on the test set as the performance metric defined by the percentage of correct segment predictions in comparison to the ground-truth segments annotated by human experts. We observe that svTCN performs better among the unsupervised metric learning approaches without using any of the action segment labels. The unsupervised metric learning approaches perform well with RNN, but other sequence models including HMMs/HSMMs and CRFs find it difficult to encode the action segments. On the other hand, Motion2Vec representation with triplet loss performs well with both the supervised and the unsupervised sequence learning approaches by better grouping the action segments with the use of labeled demonstrations. Triplet loss with RNN gives better performance among all the compared approaches. Moreover, M2V-T, a variant of Motion2Vec with a combination of labeled triplet and svTCN loss, better aligns the images temporally with nearest neighbour imitation accuracy of $84.4$\% in the embedding space.



Fig.~\ref{fig: seg-dnn-hsmm} gives a qualitative performance analysis of different sequence learning approaches on a suturing demonstration from the test set. The sequence learning models are able to predict temporally robust action segments on Motion2Vec trained embeddings. Although the video frames in the test set are not observed, Motion2Vec is able to associate them with complex segments such as needle insertion and extraction, while being invariant across camera viewpoint, background and skill level of the surgeons in the observations.

Fig.~\ref{fig: Conf_suturing} shows the confusion matrix instance of Motion2Vec with an overall evaluation segmentation accuracy of $86.07\%$ respectively. We observe that the similar neighbouring segments in the test set tend to be more often confused suggesting that the related activities are closely grouped in the embedding space.

\begin{figure}[!tbp]
\centering
\includegraphics[trim={0.1cm 0.01cm 0.1cm 0.3cm},clip,scale = 0.5]{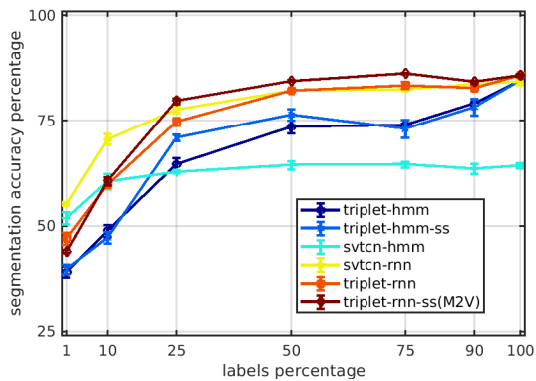}
\caption{\footnotesize Effect of percentage of labeled demonstrations on the segmentation accuracy: Motion2Vec performs better than competing unsupervised approaches using $25$\% or more labeled demonstrations. Results are averaged over $5$ iterations.} \label{fig: ss_m2v}
\end{figure}

\subsection{Effect of Labeled Examples in Semi-Supervised Learning}\label{ref: ss_exp}
We empirically investigate the effect of unlabeled training data and the relative sample size of labeled training data in Fig.~\ref{fig: ss_m2v}. As the percentage of labeled demonstrations increases in the training set from $1$ labeled demonstration to all $62$ labeled demonstrations, the segmentation accuracy of RNN with unsupervised svTCN and supervised triplet loss increases. The segmentation accuracy of HMM grows steadily with supervised triplet loss, but not with svTCN due to lack of grouped action segments. Note that for HMM, the hidden states are assigned to one of the segment labels in a greedy manner during training, and the same component to segment label map is used after Viterbi decoding on the unlabeled training demonstrations to evaluate the segmentation accuracy. Semi-supervised learning with Motion2Vec (triplet-rnn-ss) using $25$\% or more labeled demonstrations gives better segmentation accuracy over other competing approaches. The results suggest that the time-driven self-supervised embedding approaches can leverage upon a small set of labeled examples to group the action segments and semantically align the embedding space.



%
\subsection{Kinematic Imitation on da Vinci Robot}
\begin{figure}[!tbp]
\centering
\includegraphics[trim={0.6cm 0.7cm 0.3cm 0.4cm},clip,scale = 0.20]{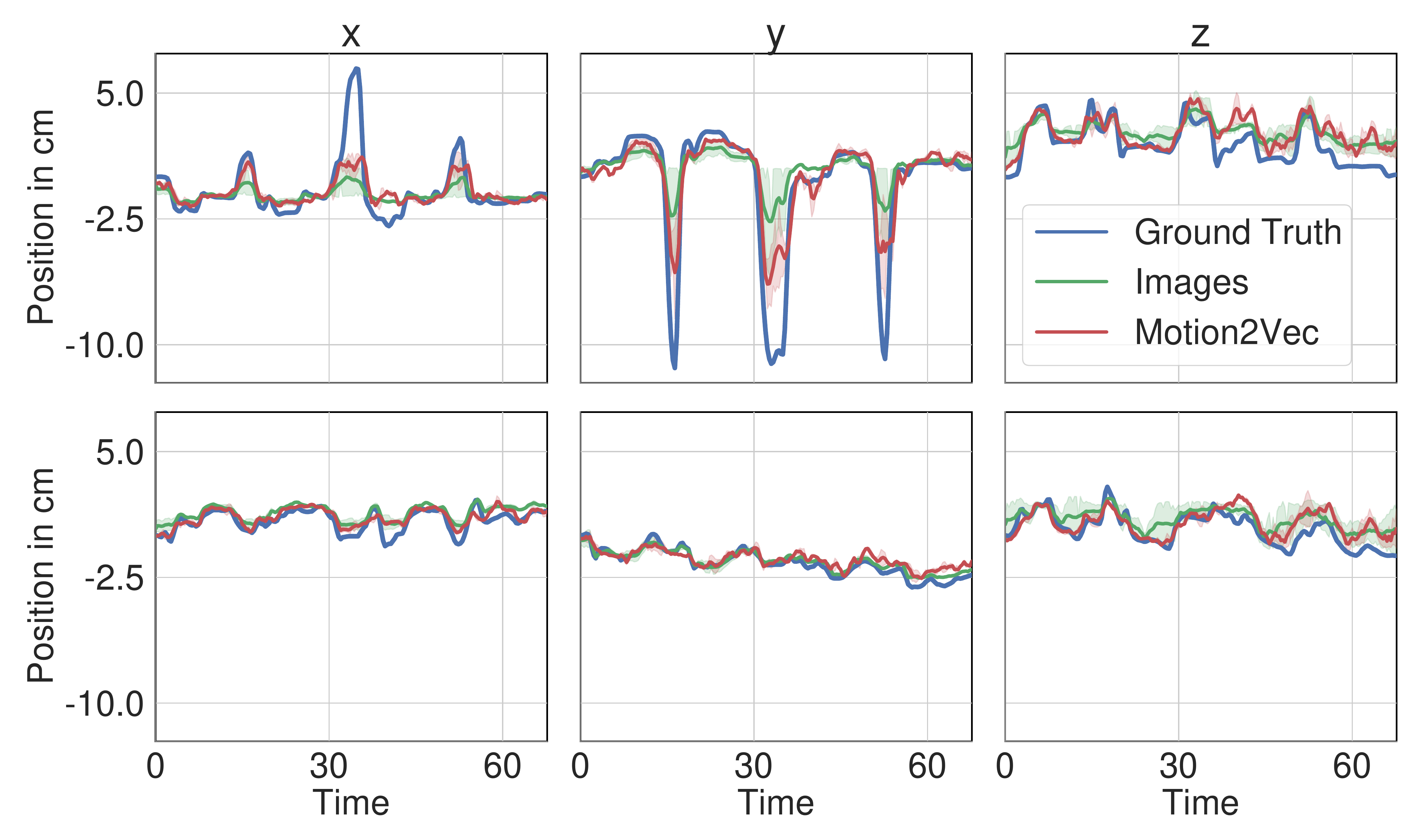}
\caption{\footnotesize Pose imitation on da Vinci robot arms from the Motion2Vec embedded video sequences using a feedforward neural network for left arm on \textit{(top)} and right arm on \textit{(bottom)} in comparison to ground-truth and image decoded poses. Results suggest $0.94$ centimeter error in position per observation on the evaluation set. Ground-truth in blue, raw videos in green, predicted in red.} \label{fig: pose_imitation}
\end{figure}

We investigate two scenarios with Motion2Vec embeddings: 1) a single pose imitation model for all surgeons, 2) a separate pose imitation model for each surgeon. Results of pose imitation from Motion2Vec in comparison to decoding from raw videos are summarized in Table~\ref{tab: pose_comp}. We get comparable performance on the test set for all surgeons with raw videos and Motion2Vec embeddings, while the per surgeon pose decoding model with Motion2Vec gives better performance with position error of $0.94$ centimeter per observation on the test set. We further test the robustness of the embeddings by adding a Gaussian noise of variance $0.15$ on top of preprocessed images and observe that Motion2Vec robustly preserves the spatio-temporal alignment of the videos. Fig.~\ref{fig: pose_imitation} shows the mean of decoded positions (with and without noise) from the embedded observations in comparison to the ground-truth and the raw videos decoded positions. We encourage the readers to see supplementary materials for imitation in simulation and real on da Vinci robot arms.

\begin{table}[tb]
\caption{\footnotesize Pose imitation error in terms of median cosine quaternion loss and root mean squared error (RMSE) in Cartesian $3$-dimensional position space in centimeters on the test set. Rows indicate pose imitation of: all surgeons from raw videos (images-all); all surgeons on Motion2Vec (M2V-all); per surgeon pose decoding on M2V (M2V-per). Motion2Vec robustly preserves the spatio-temporal alignment in the observations.} \normalsize \centering \label{tab: pose_comp}
\begin{tabular}{|c|c||c|c|}
\hline
  & \multirow{2}{*}{Noise}  & Median Cosine & RMSE \\ 
 & & Quat Loss & Position    \\ \hline \hline
\multirow{2}{*}{\textbf{Images - All}} & $-$ & $63.87$ & $1.04$ \\ 
 & $0.15$ & $89.97$ & $1.50$\\ \hline
\multirow{2}{*}{\textbf{M2V - All}} & $-$& $61.49$ & $1.15$\\ 
& $0.15$& $59.89$ & $1.34$ \\ \hline
\multirow{2}{*}{\textbf{M2V - Per}} & $-$  & $\textbf{36.49}$ & $\textbf{0.94}$\\
& $0.15$  & $\textbf{37.21}$ & $\textbf{1.12}$ \\\hline
\end{tabular}
\end{table}

\section{Conclusions and Future Work}
%



Learning from video demonstrations is a promising approach to teach manipulation skills to robots. We present Motion2Vec representation learning algorithm that groups similar action segments in a deep embedding feature space in a semi-supervised manner, improving upon the interpretability and the segmentation performance over several state-of-the-art methods. We demonstrate its use on the dual arm da Vinci robot arm to imitate surgical suturing poses from video demonstrations. In future work, we plan to learn closed loop policies on the real robot from the embedded video representations. We are also interested in providing useful feedback for training and assistance in remote surgical procedures~\cite{Tanwani_IROS_17}.

A number of directions have evolved from this work. First, there is an inherent coupling between the two arms in sub-tasks such as needle hand-off suggesting the need of bimanual coordination in the planning stage. Second, the learned disentangled representations may not generalize if there is a large difference between the videos and the real robot environment. A more feasible approach can be to learn domain invariant feature representations such as with adversarial learning~\cite{Goodfellow14}. Finally, the inherent cyclic nature of suturing task calls for learning compositional structure of the action segments. 






\section*{ACKNOWLEDGMENTS}
\footnotesize

This research was performed at the AUTOLAB at UC Berkeley in affiliation with the Berkeley AI Research (BAIR) Lab, Berkeley Deep Drive (BDD), the Real-Time Intelligent Secure Execution (RISE) Lab, the CITRIS ``People and Robots'' (CPAR) Initiative, and the NSF Scalable Collaborative Human-Robot Learning (SCHooL) Project $1734633$. The authors would like to thank Daniel Seita, Minho Hwang and our collaborators from Google, Intel and SRI for their feedback.

\bibliographystyle{IEEEtran}
\balance
\bibliography{bibliography}

\begin{thebibliography}{10}
\providecommand{\url}[1]{#1}
\csname url@samestyle\endcsname
\providecommand{\newblock}{\relax}
\providecommand{\bibinfo}[2]{#2}
\providecommand{\BIBentrySTDinterwordspacing}{\spaceskip=0pt\relax}
\providecommand{\BIBentryALTinterwordstretchfactor}{4}
\providecommand{\BIBentryALTinterwordspacing}{\spaceskip=\fontdimen2\font plus
\BIBentryALTinterwordstretchfactor\fontdimen3\font minus
  \fontdimen4\font\relax}
\providecommand{\BIBforeignlanguage}[2]{{%
\expandafter\ifx\csname l@#1\endcsname\relax
\typeout{** WARNING: IEEEtran.bst: No hyphenation pattern has been}%
\typeout{** loaded for the language `#1'. Using the pattern for}%
\typeout{** the default language instead.}%
\else
\language=\csname l@#1\endcsname
\fi
#2}}
\providecommand{\BIBdecl}{\relax}
\BIBdecl

\bibitem{Tomas_NIPS_13}
T.~Mikolov, I.~Sutskever, K.~Chen, G.~S. Corrado, and J.~Dean, ``Distributed
  representations of words and phrases and their compositionality,'' in
  \emph{Advances in Neural Information Processing Systems 26}, C.~J.~C. Burges,
  L.~Bottou, M.~Welling, Z.~Ghahramani, and K.~Q. Weinberger, Eds., 2013, pp.
  3111--3119.

\bibitem{Jang_grasp2vec_18}
\BIBentryALTinterwordspacing
E.~Jang, C.~Devin, V.~Vanhoucke, and S.~Levine, ``Grasp2vec: Learning object
  representations from self-supervised grasping,'' \emph{CoRR}, vol.
  abs/1811.06964, 2018. [Online]. Available:
  \url{http://arxiv.org/abs/1811.06964}
\BIBentrySTDinterwordspacing

\bibitem{Fang18}
K.~{Fang}, T.~{Wu}, D.~{Yang}, S.~{Savarese}, and J.~J. {Lim}, ``Demo2vec:
  Reasoning object affordances from online videos,'' in \emph{2018 IEEE/CVF
  Conference on Computer Vision and Pattern Recognition}, 2018, pp. 2139--2147.

\bibitem{Rabiner89}
L.~R. Rabiner, ``A tutorial on hidden {M}arkov models and selected applications
  in speech recognition,'' \emph{Proc. {IEEE}}, vol. 77:2, pp. 257--285, 1989.

\bibitem{Yeung16}
S.~Yeung, O.~Russakovsky, G.~Mori, and L.~Fei{-}Fei, ``End-to-end learning of
  action detection from frame glimpses in videos,'' in \emph{{IEEE} Conference
  on Computer Vision and Pattern Recognition {CVPR}}, 2016, pp. 2678--2687.

\bibitem{Zhou18}
\BIBentryALTinterwordspacing
B.~Zhou, A.~Andonian, and A.~Torralba, ``Temporal relational reasoning in
  videos,'' \emph{CoRR}, vol. abs/1711.08496, 2017. [Online]. Available:
  \url{http://arxiv.org/abs/1711.08496}
\BIBentrySTDinterwordspacing

\bibitem{Stolle02}
M.~Stolle and D.~Precup, \emph{Learning Options in Reinforcement
  Learning}.\hskip 1em plus 0.5em minus 0.4em\relax Springer Berlin Heidelberg,
  2002, pp. 212--223.

\bibitem{Kupiec92}
J.~Kupiec, ``Robust part-of-speech tagging using a hidden markov model,''
  \emph{Computer Speech and Language}, vol.~6, no.~3, pp. 225 -- 242, 1992.

\bibitem{Dwibedi_19}
\BIBentryALTinterwordspacing
D.~Dwibedi, Y.~Aytar, J.~Tompson, P.~Sermanet, and A.~Zisserman, ``Temporal
  cycle-consistency learning,'' \emph{CoRR}, vol. abs/1904.07846, 2019.
  [Online]. Available: \url{http://arxiv.org/abs/1904.07846}
\BIBentrySTDinterwordspacing

\bibitem{Sermanet17}
\BIBentryALTinterwordspacing
P.~Sermanet, C.~Lynch, J.~Hsu, and S.~Levine, ``Time-contrastive networks:
  Self-supervised learning from multi-view observation,'' \emph{CoRR}, vol.
  abs/1704.06888, 2017. [Online]. Available:
  \url{http://arxiv.org/abs/1704.06888}
\BIBentrySTDinterwordspacing

\bibitem{Lea_16}
C.~Lea, R.~Vidal, A.~Reiter, and G.~D. Hager, ``Temporal convolutional
  networks: {A} unified approach to action segmentation,'' \emph{CoRR}, vol.
  abs/1608.08242, 2016.

\bibitem{Yu10}
S.-Z. Yu, ``Hidden semi-{M}arkov models,'' \emph{Artificial Intelligence}, vol.
  174, pp. 215--243, 2010.

\bibitem{Lafferty01}
\BIBentryALTinterwordspacing
J.~D. Lafferty, A.~McCallum, and F.~C.~N. Pereira, ``Conditional random fields:
  Probabilistic models for segmenting and labeling sequence data,'' in
  \emph{Proceedings of the Eighteenth International Conference on Machine
  Learning}, ser. ICML '01.\hskip 1em plus 0.5em minus 0.4em\relax Morgan
  Kaufmann Publishers Inc., 2001, pp. 282--289. [Online]. Available:
  \url{http://dl.acm.org/citation.cfm?id=645530.655813}
\BIBentrySTDinterwordspacing

\bibitem{DiPietro_16}
R.~S. DiPietro, C.~Lea, A.~Malpani, N.~Ahmidi, S.~S. Vedula, G.~I. Lee, M.~R.
  Lee, and G.~D. Hager, ``Recognizing surgical activities with recurrent neural
  networks,'' \emph{CoRR}, vol. abs/1606.06329, 2016.

\bibitem{Ijspeert13}
A.~Ijspeert, J.~Nakanishi, P.~Pastor, H.~Hoffmann, and S.~Schaal, ``Dynamical
  movement primitives: Learning attractor models for motor behaviors,''
  \emph{Neural Computation}, no.~25, pp. 328--373, 2013.

\bibitem{Calinon17}
S.~Calinon and D.~Lee, ``Learning control,'' in \emph{Humanoid Robotics: a
  Reference}, P.~Vadakkepat and A.~Goswami, Eds.\hskip 1em plus 0.5em minus
  0.4em\relax Springer, 2018.

\bibitem{Tanwani18}
A.~K. Tanwani, ``{Generative Models for Learning Robot Manipulation Skills from
  Humans},'' Ph.D. dissertation, Ecole Polytechnique Federale de Lausanne,
  Switzerland, 2018.

\bibitem{Ho16}
\BIBentryALTinterwordspacing
J.~Ho and S.~Ermon, ``Generative adversarial imitation learning,'' \emph{CoRR},
  vol. abs/1606.03476, 2016. [Online]. Available:
  \url{http://arxiv.org/abs/1606.03476}
\BIBentrySTDinterwordspacing

\bibitem{Duan17}
\BIBentryALTinterwordspacing
Y.~Duan, M.~Andrychowicz, B.~Stadie, J.~Ho, J.~Schneider, I.~Sutskever,
  P.~Abbeel, and W.~Zaremba, ``One-shot imitation learning,'' \emph{CoRR}, vol.
  abs/1703.07326, 2017. [Online]. Available:
  \url{http://arxiv.org/abs/1703.07326}
\BIBentrySTDinterwordspacing

\bibitem{Ross10}
S.~Ross, G.~J. Gordon, and J.~A. Bagnell, ``A reduction of imitation learning
  and structured prediction to no-regret online learning,'' \emph{International
  Conference on Artificial Intelligence and Statistics}, 2011.

\bibitem{Torabi_18}
\BIBentryALTinterwordspacing
F.~Torabi, G.~Warnell, and P.~Stone, ``Behavioral cloning from observation,''
  \emph{CoRR}, vol. abs/1805.01954, 2018. [Online]. Available:
  \url{http://arxiv.org/abs/1805.01954}
\BIBentrySTDinterwordspacing

\bibitem{Osa18}
T.~Osa, J.~Pajarinen, G.~Neumann, A.~Bagnell, P.~Abbeel, and J.~Peters,
  \emph{An Algorithmic Perspective on Imitation Learning}.\hskip 1em plus 0.5em
  minus 0.4em\relax Now Publishers Inc., 2018.

\bibitem{Argall09}
B.~D. Argall, S.~Chernova, M.~Veloso, and B.~Browning, ``A survey of robot
  learning from demonstration,'' \emph{Robot. Auton. Syst.}, vol.~57, no.~5,
  pp. 469--483, May 2009.

\bibitem{zeng_3dmatch_2016}
A.~Zeng, S.~Song, M.~Nie{\ss}ner, M.~Fisher, J.~Xiao, and T.~Funkhouser,
  ``3dmatch: Learning local geometric descriptors from rgb-d reconstructions,''
  in \emph{CVPR}, 2017.

\bibitem{Schmidt_RAL_17}
T.~{Schmidt}, R.~{Newcombe}, and D.~{Fox}, ``Self-supervised visual descriptor
  learning for dense correspondence,'' \emph{IEEE Robotics and Automation
  Letters}, vol.~2, no.~2, pp. 420--427, April 2017.

\bibitem{Florence_dod_18}
P.~R. Florence, L.~Manuelli, and R.~Tedrake, ``Dense object nets: Learning
  dense visual object descriptors by and for robotic manipulation,''
  \emph{CoRR}, vol. abs/1806.08756, 2018.

\bibitem{Kuehne15}
\BIBentryALTinterwordspacing
H.~Kuehne and T.~Serre, ``Towards a generative approach to activity recognition
  and segmentation,'' \emph{CoRR}, vol. abs/1509.01947, 2015. [Online].
  Available: \url{http://arxiv.org/abs/1509.01947}
\BIBentrySTDinterwordspacing

\bibitem{Kuehne19}
\BIBentryALTinterwordspacing
H.~Kuehne, A.~Richard, and J.~Gall, ``A hybrid {RNN-HMM} approach for weakly
  supervised temporal action segmentation,'' \emph{CoRR}, vol. abs/1906.01028,
  2019. [Online]. Available: \url{http://arxiv.org/abs/1906.01028}
\BIBentrySTDinterwordspacing

\bibitem{Tang_cvpr_2012}
K.~Tang, ``Learning latent temporal structure for complex event detection,'' in
  \emph{Proceedings of the 2012 IEEE Conference on Computer Vision and Pattern
  Recognition (CVPR)}, ser. CVPR '12, 2012, pp. 1250--1257.

\bibitem{Liu17}
\BIBentryALTinterwordspacing
Y.~Liu, A.~Gupta, P.~Abbeel, and S.~Levine, ``Imitation from observation:
  Learning to imitate behaviors from raw video via context translation,''
  \emph{CoRR}, vol. abs/1707.03374, 2017. [Online]. Available:
  \url{http://arxiv.org/abs/1707.03374}
\BIBentrySTDinterwordspacing

\bibitem{Doersch15}
\BIBentryALTinterwordspacing
C.~Doersch, A.~Gupta, and A.~A. Efros, ``Unsupervised visual representation
  learning by context prediction,'' \emph{CoRR}, vol. abs/1505.05192, 2015.
  [Online]. Available: \url{http://arxiv.org/abs/1505.05192}
\BIBentrySTDinterwordspacing

\bibitem{Misra16}
\BIBentryALTinterwordspacing
I.~Misra, C.~L. Zitnick, and M.~Hebert, ``Unsupervised learning using
  sequential verification for action recognition,'' \emph{CoRR}, vol.
  abs/1603.08561, 2016. [Online]. Available:
  \url{http://arxiv.org/abs/1603.08561}
\BIBentrySTDinterwordspacing

\bibitem{Wang15a}
\BIBentryALTinterwordspacing
X.~Wang and A.~Gupta, ``Unsupervised learning of visual representations using
  videos,'' \emph{CoRR}, vol. abs/1505.00687, 2015. [Online]. Available:
  \url{http://arxiv.org/abs/1505.00687}
\BIBentrySTDinterwordspacing

\bibitem{Dwibedi_18}
\BIBentryALTinterwordspacing
D.~Dwibedi, J.~Tompson, C.~Lynch, and P.~Sermanet, ``Learning actionable
  representations from visual observations,'' \emph{CoRR}, vol. abs/1808.00928,
  2018. [Online]. Available: \url{http://arxiv.org/abs/1808.00928}
\BIBentrySTDinterwordspacing

\bibitem{Finn_VF_16}
\BIBentryALTinterwordspacing
C.~Finn and S.~Levine, ``Deep visual foresight for planning robot motion,''
  \emph{CoRR}, vol. abs/1610.00696, 2016. [Online]. Available:
  \url{http://arxiv.org/abs/1610.00696}
\BIBentrySTDinterwordspacing

\bibitem{Ebert_18}
\BIBentryALTinterwordspacing
F.~Ebert, C.~Finn, S.~Dasari, A.~Xie, A.~X. Lee, and S.~Levine, ``Visual
  foresight: Model-based deep reinforcement learning for vision-based robotic
  control,'' \emph{CoRR}, vol. abs/1812.00568, 2018. [Online]. Available:
  \url{http://arxiv.org/abs/1812.00568}
\BIBentrySTDinterwordspacing

\bibitem{Schroff15}
\BIBentryALTinterwordspacing
F.~Schroff, D.~Kalenichenko, and J.~Philbin, ``Facenet: {A} unified embedding
  for face recognition and clustering,'' \emph{CoRR}, vol. abs/1503.03832,
  2015. [Online]. Available: \url{http://arxiv.org/abs/1503.03832}
\BIBentrySTDinterwordspacing

\bibitem{Koch2015}
G.~Koch, R.~Zemel, and R.~Salakhutdinov, ``Siamese neural networks for one-shot
  image recognition,'' 2015.

\bibitem{Baraldi_15}
\BIBentryALTinterwordspacing
L.~Baraldi, C.~Grana, and R.~Cucchiara, ``A deep siamese network for scene
  detection in broadcast videos,'' \emph{CoRR}, vol. abs/1510.08893, 2015.
  [Online]. Available: \url{http://arxiv.org/abs/1510.08893}
\BIBentrySTDinterwordspacing

\bibitem{Hermans17}
\BIBentryALTinterwordspacing
A.~Hermans, L.~Beyer, and B.~Leibe, ``In defense of the triplet loss for person
  re-identification,'' \emph{CoRR}, vol. abs/1703.07737, 2017. [Online].
  Available: \url{http://arxiv.org/abs/1703.07737}
\BIBentrySTDinterwordspacing

\bibitem{Roy_18}
D.~{Roy}, C.~K. {Mohan}, and K.~S. {Rama Murty}, ``Action recognition based on
  discriminative embedding of actions using siamese networks,'' in \emph{25th
  IEEE International Conference on Image Processing (ICIP)}, 2018, pp.
  3473--3477.

\bibitem{Sen_16}
S.~{Sen}, A.~{Garg}, D.~V. {Gealy}, S.~{McKinley}, Y.~{Jen}, and K.~{Goldberg},
  ``Automating multi-throw multilateral surgical suturing with a mechanical
  needle guide and sequential convex optimization,'' in \emph{2016 IEEE
  International Conference on Robotics and Automation (ICRA)}, 2016, pp.
  4178--4185.

\bibitem{Padoy_11}
N.~{Padoy} and G.~D. {Hager}, ``Human-machine collaborative surgery using
  learned models,'' in \emph{IEEE International Conference on Robotics and
  Automation}, 2011, pp. 5285--5292.

\bibitem{Schulman_13}
J.~{Schulman}, A.~{Gupta}, S.~{Venkatesan}, M.~{Tayson-Frederick}, and
  P.~{Abbeel}, ``A case study of trajectory transfer through non-rigid
  registration for a simplified suturing scenario,'' in \emph{IEEE/RSJ
  International Conference on Intelligent Robots and Systems}, 2013, pp.
  4111--4117.

\bibitem{Lea_15}
C.~{Lea}, G.~D. {Hager}, and R.~{Vidal}, ``An improved model for segmentation
  and recognition of fine-grained activities with application to surgical
  training tasks,'' in \emph{IEEE Winter Conference on Applications of Computer
  Vision}, 2015, pp. 1123--1129.

\bibitem{Murali16}
A.~{Murali}, A.~{Garg}, S.~{Krishnan}, F.~T. {Pokorny}, P.~{Abbeel},
  T.~{Darrell}, and K.~{Goldberg}, ``Tsc-dl: Unsupervised trajectory
  segmentation of multi-modal surgical demonstrations with deep learning,'' in
  \emph{2016 IEEE International Conference on Robotics and Automation (ICRA)},
  2016, pp. 4150--4157.

\bibitem{Krishnan18}
S.~Krishnan, A.~Garg, S.~Patil, C.~Lea, G.~Hager, P.~Abbeel, and K.~Goldberg,
  \emph{Transition State Clustering: Unsupervised Surgical Trajectory
  Segmentation for Robot Learning}.\hskip 1em plus 0.5em minus 0.4em\relax
  Cham: Springer International Publishing, 2018, pp. 91--110.

\bibitem{gao14}
Y.~Gao, S.~S. Vedula, C.~E. Reiley, N.~Ahmidi, B.~Varadarajan, H.~C. Lin,
  L.~Tao, L.~Zappella, B.~B{\'e}jar, D.~D. Yuh \emph{et~al.}, ``Jhu-isi gesture
  and skill assessment working set (jigsaws): A surgical activity dataset for
  human motion modeling,'' in \emph{MICCAI Workshop}, 2014.

\bibitem{Ahmidi_17}
N.~{Ahmidi}, L.~{Tao}, S.~{Sefati}, Y.~{Gao}, C.~{Lea}, B.~B. {Haro},
  L.~{Zappella}, S.~{Khudanpur}, R.~{Vidal}, and G.~D. {Hager}, ``A dataset and
  benchmarks for segmentation and recognition of gestures in robotic surgery,''
  \emph{IEEE Transactions on Biomedical Engineering}, vol.~64, no.~9, pp.
  2025--2041, 2017.

\bibitem{Khansari11}
S.~M. Khansari-Zadeh and A.~Billard, ``Learning stable non-linear dynamical
  systems with {G}aussian mixture models,'' \emph{{IEEE} Trans. on Robotics},
  vol.~27, no.~5, pp. 943--957, 2011.

\bibitem{Tanwani_WAFR_18}
\BIBentryALTinterwordspacing
A.~K. Tanwani, J.~Lee, B.~Thananjeyan, M.~Laskey, S.~Krishnan, R.~Fox,
  K.~Goldberg, and S.~Calinon, ``Generalizing robot imitation learning with
  invariant hidden semi-markov models,'' \emph{CoRR}, vol. abs/1811.07489,
  2018. [Online]. Available: \url{http://arxiv.org/abs/1811.07489}
\BIBentrySTDinterwordspacing

\bibitem{Rippel15}
O.~Rippel, M.~Paluri, P.~Dollar, and L.~Bourdev, ``Metric learning with
  adaptive density discrimination,'' \emph{arXiv preprint arXiv:1511.05939},
  2015.

\bibitem{Zhao_06}
H.~Zhao, P.~C. Yuen, and J.~T. Kwok, ``A novel incremental principal component
  analysis and its application for face recognition,'' \emph{Trans. Sys. Man
  Cyber. Part B}, vol.~36, no.~4, pp. 873--886, 2006.

\bibitem{Sohn_16}
K.~Sohn, ``Improved deep metric learning with multi-class n-pair loss
  objective,'' in \emph{Advances in Neural Information Processing Systems 29},
  D.~D. Lee, M.~Sugiyama, U.~V. Luxburg, I.~Guyon, and R.~Garnett, Eds.\hskip
  1em plus 0.5em minus 0.4em\relax Curran Associates, Inc., 2016, pp.
  1857--1865.

\bibitem{Hochreiter_97}
\BIBentryALTinterwordspacing
S.~Hochreiter and J.~Schmidhuber, ``Long short-term memory,'' \emph{Neural
  Comput.}, vol.~9, no.~8, pp. 1735--1780, Nov. 1997. [Online]. Available:
  \url{http://dx.doi.org/10.1162/neco.1997.9.8.1735}
\BIBentrySTDinterwordspacing

\bibitem{Graves_12}
A.~Graves, \emph{{Supervised Sequence Labelling with Recurrent Neural
  Networks}}, ser. Studies in Computational Intelligence.\hskip 1em plus 0.5em
  minus 0.4em\relax Berlin: Springer, 2012.

\bibitem{Lee_13}
D.-H. Lee, ``Pseudo-label : The simple and efficient semi-supervised learning
  method for deep neural networks,'' \emph{ICML Workshop : Challenges in
  Representation Learning (WREPL)}, 07 2013.

\bibitem{Sutton12}
\BIBentryALTinterwordspacing
C.~Sutton and A.~McCallum, ``An introduction to conditional random fields,''
  \emph{Found. Trends Mach. Learn.}, vol.~4, no.~4, pp. 267--373, Apr. 2012.
  [Online]. Available: \url{http://dx.doi.org/10.1561/2200000013}
\BIBentrySTDinterwordspacing

\bibitem{Vakanski12}
A.~Vakanski, I.~Mantegh, A.~Irish, and F.~Janabi-Sharifi, ``Trajectory learning
  for robot programming by demonstration using hidden markov model and dynamic
  time warping,'' \emph{IEEE Transactions on Systems, Man, and Cybernetics,
  Part B (Cybernetics)}, vol.~42, no.~4, pp. 1039--1052, 2012.

\bibitem{Tanwani16}
A.~K. Tanwani and S.~Calinon, ``Learning robot manipulation tasks with
  task-parameterized semitied hidden semi-markov model,'' \emph{IEEE Robotics
  and Automation Letters}, vol.~1, no.~1, pp. 235--242, 2016.

\bibitem{Finn_ss_15}
\BIBentryALTinterwordspacing
C.~Finn, X.~Y. Tan, Y.~Duan, T.~Darrell, S.~Levine, and P.~Abbeel, ``Learning
  visual feature spaces for robotic manipulation with deep spatial
  autoencoders,'' \emph{CoRR}, vol. abs/1509.06113, 2015. [Online]. Available:
  \url{http://arxiv.org/abs/1509.06113}
\BIBentrySTDinterwordspacing

\bibitem{Tanwani_IROS_17}
A.~K. Tanwani and S.~Calinon, ``A generative model for intention recognition
  and manipulation assistance in teleoperation,'' in \emph{{IEEE/RSJ}
  International Conference on Intelligent Robots and Systems, {IROS}}, 2017,
  pp. 43--50.

\bibitem{Goodfellow14}
\BIBentryALTinterwordspacing
I.~Goodfellow, J.~Pouget-Abadie, M.~Mirza, B.~Xu, D.~Warde-Farley, S.~Ozair,
  A.~Courville, and Y.~Bengio, ``{Generative Adversarial Nets},'' in
  \emph{Advances in Neural Information Processing Systems 27}, Z.~Ghahramani,
  M.~Welling, C.~Cortes, N.~D. Lawrence, and K.~Q. Weinberger, Eds., 2014, pp.
  2672--2680. [Online]. Available:
  \url{http://papers.nips.cc/paper/5423-generative-adversarial-nets.pdf}
\BIBentrySTDinterwordspacing

\end{thebibliography}

\end{document}